\pdfoutput=1

\documentclass[11pt]{article}

\usepackage{emnlp2021}

\usepackage{times}
\usepackage{latexsym}

\usepackage[T1]{fontenc}

\usepackage[utf8]{inputenc}

\usepackage{microtype}

\usepackage{amsmath,amsfonts,bm}

\def\eqref#1{equation~\ref{#1}}

\def\1{\bm{1}}

\DeclareMathAlphabet{\mathsfit}{\encodingdefault}{\sfdefault}{m}{sl}
\SetMathAlphabet{\mathsfit}{bold}{\encodingdefault}{\sfdefault}{bx}{n}

\DeclareMathOperator*{\argmin}{arg\,min}

\usepackage{subfiles}
\usepackage{url}
\usepackage{caption}
\usepackage{graphicx}
\usepackage{amsfonts}
\usepackage{amsmath}
\usepackage{float}
\usepackage{amsthm}
\usepackage{subfigure}
\usepackage{amssymb}
\usepackage{mathtools}
\usepackage{bbm}
\usepackage{multirow}

\theoremstyle{definition}
\newtheorem{example}{Example}
\newtheorem{hypothesis}{Hypothesis}

 \usepackage{xcolor}

\title{Exposure Bias versus Self-Recovery: \\ Are Distortions Really Incremental for Autoregressive Text Generation?}

\author{Tianxing He \\
  MIT \\
  \texttt{cloudygoose@csail.mit.edu} \\\And
  Jingzhao Zhang \\
  MIT \\
  \texttt{jzhzhang@mit.edu} \\\AND
  Zhiming Zhou \\
  Shanghai Jiao Tong University \\
  \texttt{heyohai@apex.sjtu.edu.cn} \\ \And
  James Glass \\
  MIT\\
  \texttt{glass@mit.edu} \\
  }

\begin{document}
\maketitle
\begin{abstract}
Exposure bias has been regarded as a central problem for auto-regressive language models (LM). It claims that teacher forcing would cause the test-time generation to be incrementally distorted due to the training-generation discrepancy.  Although a lot of algorithms have been proposed to avoid teacher forcing and therefore alleviate exposure bias, there is little work showing how serious the exposure bias problem actually is. In this work, we focus on the task of open-ended language generation, propose metrics to quantify the impact of exposure bias in the aspects of quality, diversity, and consistency. Our key intuition is that if we feed ground-truth data prefixes (instead of prefixes generated by the model itself) into the model and ask it to continue the generation, the performance should become much better because the training-generation discrepancy in the prefix is removed. Both automatic and human evaluations are conducted in our experiments. On the contrary to the popular belief in exposure bias, we find that the the distortion induced by the prefix discrepancy is limited, and does not seem to be incremental during the generation. Moreover, our analysis reveals an interesting self-recovery ability of the LM, which we hypothesize to be countering the harmful effects from exposure bias.
\end{abstract}

\section{Introduction}
\label{sec:intro}

Language model (LM) has been a central module for natural language generation (NLG) tasks \citep{trends-nlp} such as open-ended language generation \citep{radford18gpt2, gpt32020brown}, machine translation \citep{wu17ganmt}, dialogue response generation \citep{dialogue17jiwei}, image captioning \citep{coco14tsung}, etc. For decades, maximum likelihood estimation (MLE) has been the most widely used objective for LM training. However, there is a popular belief in the natural language processing (NLP) community that standard MLE training 
suffers from the \textit{exposure bias} problem which leads to an incremental performance degradation during test-time generation.

The claim of the exposure bias problem \citep{ss15bengio,seqtrainrnn16marc} is originated from the following discrepancy between MLE training and test-time generation for auto-regressive language models: During training, the model is trained to predict the next word conditioned on prefix (or history) words sampled from the ground-truth data distribution; While during generation, the model generates words conditioned on prefix sequences sampled from the model itself. Due to the \textit{exposure} to real data during training, the language model could potentially be \textit{biased} to only perform well with data prefixes. Therefore, it is claimed (and widely believed among researchers) that during generation the errors could accumulate along the generated sequence, and the distribution generated by the model would be \textbf{incrementally distorted}. The forced exposure to ground-truth data during training is also referred to as \textit{teacher forcing}. 

\begin{figure}
  \centering
  \includegraphics[width=0.8\linewidth]{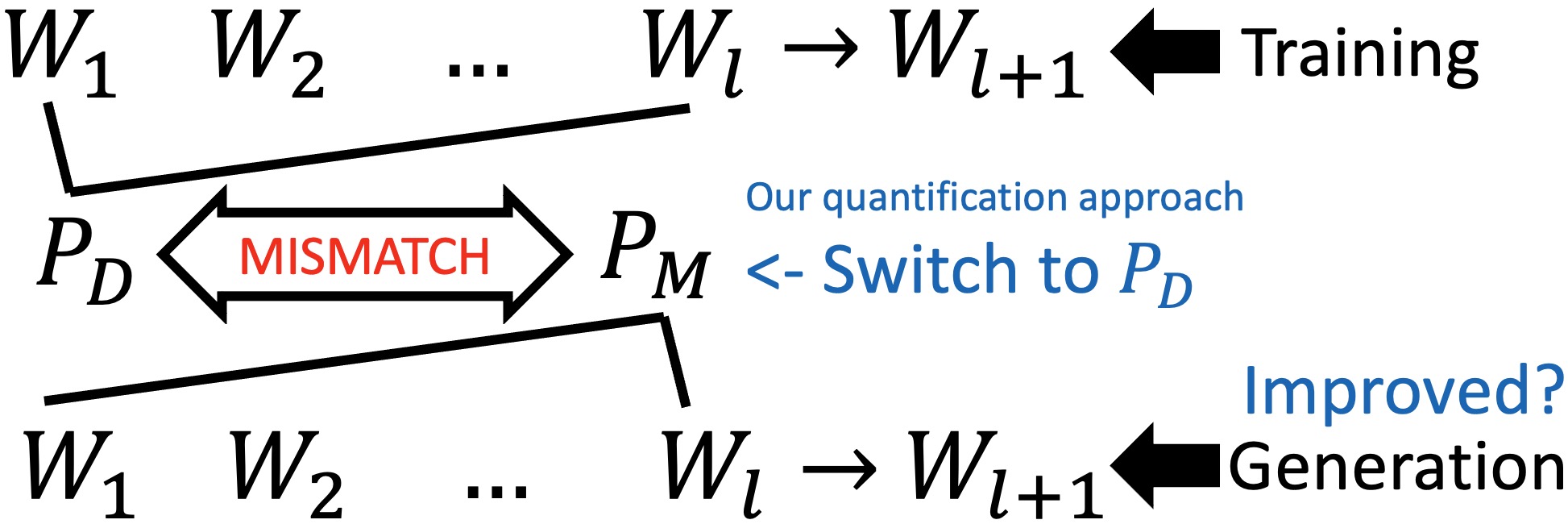}
  \caption{An illustration of the training-generation discrepancy and our key intuition for verifying exposure bias (in blue): During generation, if we feed ground-truth data prefixes (from $P_D$) instead of the model's own samples ($P_M$), the generation performance should be better because the prefix discrepancy is removed. }
  \label{fig:eb_intuition}
\end{figure}



To avoid teacher forcing, many training algorithms \citep{ss15bengio,Lamb2016ProfessorFA,seqtrainrnn16marc,yu2016seqgan,zhu2018texygen,cot18sidi,rankgan17,leakgan17jiaxian,advtext17sai,seq2seqbeam16sam,nie2018relgan,zhan18irlgan, scratchgan,DBLP:journals/corr/RennieMMRG16,feng-etal-2021-guiding} have been proposed as alternatives to MLE training for text generation. Many of these works utilize techniques from generative adversarial networks (GANs) \citep{Goodfellow14gan} or reinforcement learning (RL) \citep{sutton98rl}. In  this work, we refer to these algorithms as non-MLE methods or text GANs.

With the huge research efforts devoted to alleviate exposure bias, interestingly, the existence or significance of exposure bias is much less studied. Interestingly, despite the criticism, MLE (teacher forcing) has remained to be the dominant objective for LM training \citep{radford18gpt2,keskar2019ctrl,gpt32020brown}. On the other hand, non-MLE training methods are still struggling to outperform the MLE baseline \citep{shortgan18massimo, scratchgan}. These developments lead us to question: \textit{Is exposure bias truly a serious problem for MLE training?}

In this work we seek a direct answer to the above question. Here we briefly summarize our contributions: We design and experiment with various metrics to quantify the impact from exposure bias. On the contrary to our expectation, our measurements consistently show that removing the prefix discrepancy only brings limited gain, and the incremental distortion is not observed. Moreover, our analysis reveals an interesting self-recovery ability of the LM, which we hypothesize to be countering the harmful effects from exposure bias.

\section{Related Works}
\label{sec:related}

We first clarify that ``Whether exposure bias is serious for MLE training?" and ``Whether new non-MLE algorithms improve generation performance?" are two related but different questions, and our work has a clear focus on the first question. Despite the large body of works (listed in Section \ref{sec:intro}) devoted to alleviate exposure bias, its actual impact has rarely been systematically studied or validated in a direct and principled way. \citet{ebmt19wen} attempts to measure the gain from alleviating exposure bias, by counting the ground truth words whose probabilities in the predicted distributions produced by their proposed model are greater than those from the baseline model. However, it is unclear how this experiment can be linked to exposure bias. \citet{schmidt2019generalization} provides valuable discussions around the claim of exposure bias, but they did not propose an operational definition or metric. \citet{yifan19rethinkeb} attempts to measure exposure bias by comparing the model's performance on seen (training) data and unseen (test) data. However, this methodology is more about the generalization gap, instead of exposure bias. Finally, \citet{wang2020exposure} discusses the potential link between exposure bias and hallucination in the context of machine translation.

In a relevant direction to answer the second question, several recent works attempt to evaluate whether the non-MLE training methods can really give superior NLG performance than standard MLE training for open-ended text generation.  \citet{shortgan18massimo} tunes a ``temperature'' parameter in the softmax output, and evaluates models over the whole quality-diversity spectrum. \citet{evalgan18stanislau} proposes to use ``Reverse Language Model score'' or ``Frechet InferSent Distance'' to evaluate the model's generation performance. \citet{evalgan18guy} proposes a method for approximating a distribution over tokens from GANs, and then evaluates models with standard LM metrics. 

These works arrive at a similar conclusion: The generation performance of text GANs is not convincingly better, or even worse, than standard MLE training. These negative results motivate us to reassess the exposure bias problem, which serves as a major motivation of text GANs. In the next section, we begin by introducing notations and background.


\begin{table*}[t]
\small
    \centering
    
    \begin{tabular}{l}
    \hline
    \texttt{<LEN-20 PROMPT>} ... \texttt{(<LEN-100 \textbf{DATA} PREFIX>)} ... in February 1942 , Headlam $\downarrow$ \\
    \textbf{Generation:} was captain of No. 4 Squadron , which had recently been withdrawn from service \\ in France .  He was promoted to squadron leader in November that year , and appointed ... \\
    \hline
    \texttt{<LEN-20 PROMPT>} ... \texttt{(<LEN-100 \textbf{MODEL} PREFIX>)} ... when he was  $\downarrow$ \\
    \textbf{Generation:} appointed chief flying instructor at Moth City . He flew both jet aircraft and \\ Mitsubishi A6M Zero fighters in large numbers during the weeks following the war , when he ... \\
    \hline
    ... \texttt{(<LEN-120 \textbf{SHUFFLED DATA} PREFIX>)} ... ( Air Marshal  against West April July $\downarrow$ \\
    \textbf{Generation:} ) . Wing Commander WS Marais , No. 3 Wing ( Wing Commander ) replaced \\ Headlam in October 1942 . The Wing scored 9 victories in the three months between ... \\
    \hline
    ... \texttt{(<LEN-120 \textbf{RANDOM} PREFIX>)} ... Canterbury Oxford flagship looks person $\downarrow$ \\
    \textbf{Generation:} in New Canterbury City thousands more . " Yap State Party vice president Datuk \\ Bambang Ishii had died that same day of mild cancer of the left foot in ... \\
    \hline
    \end{tabular}
    \vspace{-0.2cm}
    \caption{Samples of a MLE-trained transformer LM when fed with different types of prefixes. The prompt is \textit{``= Frank Headlam = Air Vice Marshal Frank Headlam , CB , CBE ( 15 July 1914 ''}, which is the beginning of an article in the wiki-103 test set. To save space, we omit the long prefix and only show the last few words. A complete set of samples are included in Table \ref{tab:sampleinspection_transformer} (Appendix \ref{app:auxiliary_res_plots}). The examples are not cherry-picked. }
    \vspace{-0.2cm}
    \label{tab:sampleinspection}
\end{table*}

\section{Background and Notations}
\label{sec:preliminaries}

Our experiments will be focused on the task of open-ended language generation, which is arguably a good test bed for exposure bias due to the following reasons: (1) The generation length is long. (2) Different from typical seq2seq tasks such as machine translation, the generation space is only weakly constrained and the topics can be very diverse, which means the training-generation discrepancy could be large.

\paragraph{Notations} Auto-regressive language models are trained to learn the probability distribution of the $(l+1)_{\text{th}}$ word (or token) $W_{l+1}$ in a sentence $W$ conditioned on the prefix $W_{1:l} := (W_1, \dots, W_{l})$ and prompt $C$. We use $W_i \in V$ to denote a discrete random variable distributed across the vocabulary $V$. For simplicity, we assume all sentences are of length $L$ in the formulations. Denoting the ground-truth data distribution as $P_{D}$, the standard MLE (also referred to as teacher forcing) training aims to minimize the negative log-likelihood (NLL) below:
\begin{align}
\small
\label{eq:mleobj}
\begin{split}
 \mathcal{L}_{\text{MLE}}= \mathop{\mathbb{E}}_{(C,W) \sim P_D}-\Sigma_{l=0}^{L-1}\log P_{\theta}(W_{l+1}|\ C,W_{1:l}),
\end{split}
\end{align}
where $P_\theta(\,\cdot\ |\ C,W_{1:l})$ denotes the conditional distribution of $W_{l+1}$ of $P_\theta$ given a prompt $C$ and a prefix $W_{1:l}$, and $\theta$ stands for the set of parameters to be trained. Note that the concept of ``sentence'' ($W$) can be naturally generalized to paragraphs or even articles, depending on the target task.

We denote the distribution of a MLE-trained LM as $P_{M}$, which is the major subject of this study. We will experiment with two popular model architectures: LSTM LM \citep{lstm-hochreiter1997long, lstmlm-speech} and transformer LM \citep{NIPS2017_3f5ee243,transformerxl19zihang}. For generation, we mainly focus on classical ancestral sampling, due to the following reasons: (1) The sampling algorithms (e.g., top-$k$ sampling) are known to trade quality out of diversity \citep{shortgan18massimo,zhang2020trading,nadeem2020systematic}. So, invoking them could ``hide'' the exposure bias problem because the prefixes from the model will be of higher quality (thus narrowing the discrepancy). (2) The sampling algorithms requires tuning of hyper-parameters, which will complicate the comparison. With these being said, in our experiments (including human evaluation) we have also tested with top-$k$ sampling \citep{fan-etal-2018-hierarchical}, with consistent observations.

In addition to popular metrics in natural language generation (NLG) such as BLEU \citep{papineni-etal-2002-bleu} or METEOR \citep{denkowski:lavie:meteor-wmt:2014}, our quantification approaches  also rely on the measurements of the divergence between two distributions.  Let $\mathcal{P}$ denote the set of probability distributions on the vocabulary $V$, and let $f_\text{div}:\mathcal{P} \times \mathcal{P} \to \mathbb{R}_{\ge 0}$ be a divergence function between two distributions. We will adopt two popular probability divergence functions: total variation distance (denoted as $d_\text{TV}$) and Jensen-Shannon divergence ($d_\text{JS}$). Definitions of $d_\text{TV}$ and $d_\text{JS}$ are provided in Appendix \ref{appsec:appmetricdef}.



\section{A Qualitative Attempt}
\label{sec:meth}
We begin with a qualitative attempt to verify the seriousness of exposure bias, by designing a \textit{prefix-switching} experiment as follows: We feed a MLE-trained transformer LM on the wiki-103 dataset  with four types of prefixes of the same length: (1) test-data samples, (2) model's own samples, (3) test-data samples shuffled on word-level, or (4) samples from a uniformly random distribution on $V$. Then we let the model continue the generation given these prefixes and compare the quality of the samples in a qualitative manner. Details of the model and dataset are deferred to Section \ref{sec:exp_all}.

The intuition behind the prefix-switching experiment follows immediately from the original claim of exposure bias: During generation, if we set the prefix distribution to be the ground-truth data distribution instead of the model's own,  the discrepancy between training and generation in the prefix will be removed, and hence the model's generation quality should be much better. We illustrate this idea in Figure \ref{fig:eb_intuition}. In the extreme case of shuffled or random prefixes, due to the claim from exposure bias about the incremental distortion, we expect the model to generate also badly distorted sequences. 

The samples with different types of prefixes are shown in Table \ref{tab:sampleinspection}. To make the generation from data and model prefix more comparable, we force the same prompt at the beginning to constrain the topic. Moreover, we intentionally use long prefixes of length 100, in the hope that the incremental distortion of generation (as claimed by exposure bias) would become observable. We also include examples from a LSTM LM in Table \ref{tab:sampleinspection_lstm} (Appendix \ref{app:auxiliary_res_plots}), which gives similar observations.

On the contrary to our expectation, we do not observe a noticeable difference in sample quality comparing samples from model and data prefixes. More surprisingly, the model is still able to generate fairly good samples from shuffled prefixes. Even in the extreme case of random prefixes, we still observe basic language structures in the sample.

This experiment suggests that the MLE-trained LMs have the \textit{self-recovery ability}, i.e., the model is able to recover from artificially distorted history input, and generate samples with reasonable quality. This observation casts doubt on exposure bias, which claims that the distortions in the generation should, on the contrary, be \textit{incremental}.

This qualitative attempt suggests the impact from exposure bias could be more subtle than we expected. Especially, it would be difficult to judge whether the distortions are incremental via qualitative examination. Motivated by this, we now turn to more rigorous quantification methods to measure the impact of exposure bias.


\section{Quantification Methods}


We now introduce the definitions of our proposed metrics. We attempt to quantify the impact of exposure bias on three key aspects of open-ended language generation: quality, diversity, and consistency. We first  introduce EB-M, which covers the quality and diversity aspects, and then EB-C, which covers the consistency aspect.
Following the intuition of the prefix switching experiment, we design our quantification metrics to be a simple ratio reflecting the relative performance gain when data prefixes are fed to the model as opposed to the original model prefixes. 

\subsection{Definition of EB-M}
\label{sec:eb_seq_formulate}

In this section, we propose the EB-M metric. Since the key idea is to compare the generation quality with different types of prefixes, denoting the prefix distribution as $P_H \in \{P_M, P_D\}$ (model or data prefixes), we first formalize the following 3-step generation process: (1) Sample a fixed-length prompt $C$ from $P_D$. (2)
    Given a prefix length $l$ and a prefix distribution $P_H$, we sample a prefix $W_{1:l}$ from $P_H(\,\cdot \ |\ C)$. (3) Conditioned on the prompt and prefix, we sample $W_{l+1:l+l_\text{gen}}$ from $P_M(\,\cdot \ |\ C,W_{1:l})$, where $l_\text{gen}$ is the length of generation. 
    

We denote the marginal distribution of $W_{l+1:l+l_\text{gen}}$ of the above generation process as $P_{M|H}^{W_{l+1:l+l_\text{gen}}}$. From the claim of exposure bias, we expect the quality or diversity of $W_{l+1:l+l_\text{gen}}$ to be better when $P_D$ is used as $P_H$ than $P_M$. 




With these ingredients in hand, we now propose the EB-M quantification metric for exposure bias (``M'' stands for ``marginal''). It reflects the relative performance gain when the length-$l$ prefix is from $P_D$ instead of from $P_M$, and is formulated as below:
\begin{equation}\label{eq:eb-bleu}
\small
    \text{EB-M}(M,l,f_\text{score}) = \frac{f_\text{score}(P_{M|D}^{W_{l+1:l+l_\text{gen}}}, P_{D}^{W_{l+1:l+l_\text{gen}}})}{f_\text{score}(P_{M|M}^{W_{l+1:l+l_\text{gen}}}, P_{D}^{W_{l+1:l+l_\text{gen}}})}.
\end{equation}
$f_\text{score}$ is a pre-defined scoring function\footnote{We include $P_D$ into the input of $f_\text{score}$, because some of the metrics (e.g., BLEU) require data samples as reference.} of the generation samples, and we assume higher value of $f_\text{score}$ indicates that the generation is of higher quality or diversity. In our experiments, we will use popular NLG metrics including BLEU \citep{papineni-etal-2002-bleu} / Nist \citep{doddington02nist} / METEOR \citep{denkowski:lavie:meteor-wmt:2014}, which mainly capture the quality aspect, and backward-BLEU \citep{zhan18irlgan} / n-gram entropy \citep{yizhe18gan_aim}, which capture the diversity aspect. 




EB-M has several potential weaknesses: First, it doesn't reflect how the generation is consistent with the given prefix $W_{1:l}$, because it only focuses on the marginal distribution of $W_{l+1:l+l_\text{gen}}$. Second, even in the $P_{M|D}$ case, exposure bias still affects the generation $W_{l+1:l+l_\text{gen}}$.\footnote{To alleviate this issue, in our experiments we will use a relatively small $l_\text{gen}$ (e.g., 20 or 10), and set prefix length $l$ to be multiple times larger than it.} To cover these shortcomings, in the next section we propose another quantification method named EB-C, which focuses on the model's word-level conditional generation distribution of $W_{l+1}$ given prefix $W_{1:l}$.  Finally, the standard NLG metrics (such as BLEU) have recently been criticized that they may correlate poorly with human judgements \citep{sellam-etal-2020-bleurt}. Therefore, in Section \ref{sec:humaneval} we conduct a human evaluation for the completeness of our evaluation. 


\subsection{Definition of EB-C}
\label{sec:ebc_def}

We propose EB-C as a \textit{conditional} counterpart of EB-M. Again, let $P_H \in \{P_M, P_D\}$ denote the prefix distribution. With a given prefix length $l$, we first define the \textit{conditional generation deviation} (CGD) as the expected distance between $P_M$ and $P_D$ for $W_{l+1}$, conditioned on the prefix samples from $P_H$, measured by divergence $f_\text{div}$: 
\begin{align}
\small
\label{eq:cgd}
\begin{split}
    &\text{CGD}({M|H(l)}, f_\text{div}) = \\
    &\mathop{\mathbb{E}}_{C \sim P_D, W_{1:l} \sim P_{H}(\cdot|C)}[f_\text{div}(P_{M}(\cdot|C, W_{1:l}), P_{D}(\cdot|C, W_{1:l}))].
\end{split}
\end{align}
For the choice of $f_\text{div}$, we will use the standard $d_\text{TV}$ and $d_\text{JS}$ divergence introduced in Section \ref{sec:preliminaries}. A smaller CGD value indicates a better-modeled conditional distribution for $W_{l+1}$ under prefix distribution $P_H$, which captures the consistency aspect of text generation. Also note that since here we focus on the generation of a \textit{single} word $W_{l+1}$, the distortion from exposure bias should be completely removed when data prefix is fed (i.e., in the $\text{CGD}(M|D)$ case).



Exposure bias should induce a meaningful gap between $\text{CGD}(M|M(l), f_\text{div})$ and $\text{CGD}(M|D(l), f_\text{div})$. We now define the EB-C quantification metric for exposure bias at prefix length $l$ with divergence $f_\text{div}$ to be:
\begin{equation}
\label{eq:eb-c-main}
    \text{EB-C}(M, l, f_\text{div})=\tfrac{\text{CGD}(M|M(l), f_\text{div})}{\text{CGD}(M|D(l), f_\text{div})}.
\end{equation}
EB-C reflects the relative gain in CGD value when the prefix distribution is replaced by $P_D$ from $P_M$. Since the computation of CGD requires access to the ground-truth data distribution, in our experiments we will first consider a synthetic setting, where an existing model is used as $P_D$.

\begin{table*}[th]
\small
\addtolength{\tabcolsep}{-1.2pt}
\centering
\begin{tabular}{c|cccccc}
\hline
\multirow{2}{*}{\bf EB-M / EB-C}                          & \multicolumn{6}{c}{\textbf{prefix length ($l$)}}         \\ \cline{2-7}
                         & \textbf{20}                & \textbf{30}                & \textbf{40}                & \textbf{50}                & \textbf{60}      & \textbf{100}          \\ \hline
{\bf EB-M} ($M_{\text{TF}}$, Nist)     & 1.000 $\pm$ .002  & 1.000 $\pm$ .001 & 1.000 $\pm$ .001  & 0.999 $\pm$ .001 & 1.000 $\pm$ .001  & 0.998 $\pm$ .001 \\
{\bf EB-M} ($M_{\text{TF}}$, METEOR)     & 1.003 $\pm$ .002  & 1.002 $\pm$ .002  & 1.003 $\pm$ .003  & 1.003 $\pm$ .002 & 1.003 $\pm$ .002  & 1.004 $\pm$ .003 \\
{\bf EB-M} ($M_{\text{TF}}$, BLEU)         & 0.998 $\pm$ .005  & 1.000 $\pm$ .004  & 1.003 $\pm$ .003  & 0.998 $\pm$ .006 & 1.001 $\pm$ .003 & 1.003 $\pm$ .005 \\
{\bf EB-M} ($M_{\text{TF}}$, back-BLEU)     & 1.002 $\pm$ .004  & 1.006 $\pm$ .002  & 1.005 $\pm$ .002  & 1.006 $\pm$ .002 & 1.008 $\pm$ .002 & 1.010 $\pm$ .003 \\
{\bf EB-M} ($M_{\text{TF}}$, entropy)     & 1.000 $\pm$ .001  & 1.000 $\pm$ .001  & 1.000 $\pm$ .001  & 1.000 $\pm$ .001 & 1.000 $\pm$ .001  & 1.000 $\pm$ .001 \\
\hline
\textbf{EB-C}$(M,d_\text{TV})$          & 1.005 $\pm$ .008 & 1.010 $\pm$ .009 & 1.001 $\pm$ .006 & 1.002 $\pm$ .008 & 1.002 $\pm$ .007 &  1.000 $\pm$ .009 \\
\textbf{EB-C}$(M, d_\text{JS})$          & 1.019 $\pm$ .011 & 1.027 $\pm$ .013 & 1.013 $\pm$ .010 & 1.013 $\pm$ .012 & 1.011 $\pm$ .011 & 1.009 $\pm$ .013 \\
\hline 
{\bf EB-M} ($M_{\text{TF}}|D_{\text{shuf}}$, BLEU)     & 1.059 $\pm$ .005 & 1.070 $\pm$ .006  & 1.082 $\pm$ .005 & 1.093 $\pm$ .006  & 1.102 $\pm$ .008  & 1.112 $\pm$ .005 \\
\textbf{EB-C}$(M|D_\text{shuf}, d_\text{JS})$          & 1.885 $\pm$ .015 & 1.886 $\pm$ .021 & 1.869 $\pm$ .019 & 1.869 $\pm$ .017 & 1.886 $\pm$ .015 & 1.926 $\pm$ .014 \\
\hline
\end{tabular}
\vspace{-0.2cm}
\caption{\label{tab:main_eb_res} EB-M and EB-C measurements of the transformer LM. $M_\text{TF}$ refers to the transformer model. To save space, we defer the corresponding $f_\text{score}$ / CGD values to Table \ref{tab:eb_m_fscore} and Table \ref{tab:eb_c_cgd} in Appendix \ref{app:auxiliary_res_plots}. Both EB-M and EB-C measurements indicate that removing the prefix discrepancy gives around 1\% or 2\% of relative performance gain. }
\vspace{-0.2cm}
\end{table*}

\section{Experiment Results}
\label{sec:exp_all}

In this section our quantification results for exposure bias are presented. In addition, we propose variants of the EB-M / EB-C metrics to analyze the self-recovery ability.

For a systematic assessment of exposure bias, we decompose the claim of exposure bias into two factors: (1) The discrepancy in the prefix distribution would hurt the generation performance, in general. (2) Moreover, the distortion should be \textit{incremental} along the generation. The first factor can be reflected by the average magnitude of the measurements (the values are expected to be larger than 1 by a meaningful margin), and the second factor can be reflected by whether the measurements have an increasing trend along the prefix length. Below we begin by describing the experiment setting.


\subsection{Experiment Setting}
Most of our experiments are conducted on the wiki-103 dataset.\footnote{\href{https://blog.einstein.ai/the-wikitext-long-term-dependency-language-modeling-dataset/}{link to the wiki-103 dataset}.} It has around 1.8m sentences / 101m words for training, and 4k sentences / 241k words for testing. We favour the wiki-103 dataset because it has long and complex paragraphs (from Wikipedia), which is useful for the measurements of exposure bias. It is also among the most popular datasets for LM benchmarking. 

\vspace{-0.1cm}
\paragraph{Real-data Setting for EB-M} To prepare a MLE-trained $P_M$, we use the code from Transformer-XL \citep{transformerxl19zihang} to train a transformer LM on the wiki-103 dataset. The model is a 16-layer Transformer-XL model with a hidden dimension of 410 and an inner dimension of 2100. Since the computation of BLEU / METEOR / Nist scores requires large amounts of unseen real-data samples as references, we use half of the wiki-103 training data (around 900k sentences and 50m words) to train the model $P_M$, and save the other half as samples from $P_D$ (used as reference for BLEU / METEOR / Nist). Other training configurations of transformer-XL (learning rate, batch size, etc.) are not changed. The resulting model $P_M$ has a test-set perplexity (PPL) of 27.81 (If trained on full training data, the PPL will be 24.02). In addition, we also train a 3-layer LSTM LM \citep{lstmlm-speech} with a hidden layer dimension of 600. It has a test-set PPL of 34.80.

\vspace{-0.1cm}
\paragraph{Synthetic Setting for EB-C} Recall that the estimation of EB-C (Equation \ref{eq:eb-c-main}) requires inference of the data distribution, therefore we first consider a synthetic setting where we treat the 16-layer transformer-XL model trained on wiki-103 full training data as $P_D$. Then, we construct a pseudo training set which is roughly the same size of the original training set by sampling from it. Next, a randomly initialized 4-layer transformer-XL model is used as $P_M$ and trained on the pseudo training set with the same hyper-parameters. The resulting $P_M$ model has a perplexity of 84 on wiki-103 test set (while $P_D$ has a perplexity of 24), indicating that the pseudo $P_D$ model is far from being fully recovered by the training process. Finally, EB-C is estimated using 10k samples from $P_M$ and $P_D$.  

In our experiments for both EB-M and EB-C, we fix the length of prompt $C$ and $l_\text{gen}$ (for EB-M) to be 20, and vary the prefix length $l$.

\vspace{-0.1cm}
\subsection{Quantification of Exposure Bias}
\label{sec:eb_exp_main}
\vspace{-0.1cm}

We show the EB-M (in real-data setting) and EB-C (in a synthetic setting) measurements with different prefix length $l$ in the upper and middle part of Table \ref{tab:main_eb_res}. The mean and standard deviation of measurements from 10 runs with different random seeds are shown as error bar. For each run, 10k samples from the model are used to calculate $f_\text{score}$ / CGD values, and for NLG metrics 10k data samples as references. The 3-gram BLEU (or Nist / entropy) score is adopted.\footnote{We have also experimented with BLEU-4 or BLEU-2, and the measurements are very similar.} To save space, we defer the EB-M measurements for the LSTM model with different metrics to Table \ref{tab:app_corpusbleu_lstm} (Appendix \ref{app:auxiliary_res_plots}), and the observations are similar.

We observe that the EB-M measurements with various NLG metrics are around or less than 1.01, for both the LSTM or transformer model. Likewise, the EB-C measurements are around or less than 1.02. While this suggests that the prefix discrepancy does hurt the generation performance to some level (1\% or 2\% of relative degradation), its impact seems to be limited. More importantly, the performance gap does not become much larger as the prefix length grows (even if we use a long prefix length of 100), which contradicts the incremental distortion claim of exposure bias.

In addition, we test several natural variants of the EB-M experiments: (1) Set $l_\text{gen}$ to be shorter (10) or longer (30); (2) Use a smaller number of references (e.g., 1k); (3) Use top-$k$ sampling with $k=40$ instead of sampling from the whole vocabulary. In all these cases the observations are very similar (the measurements are less than or around 1.01), and we omit these results here.

\begin{figure}
  \centering
  \subfigure[EB-M$(M_\text{TF}|D_\text{corrupt})$]{\includegraphics[width=0.49\linewidth]{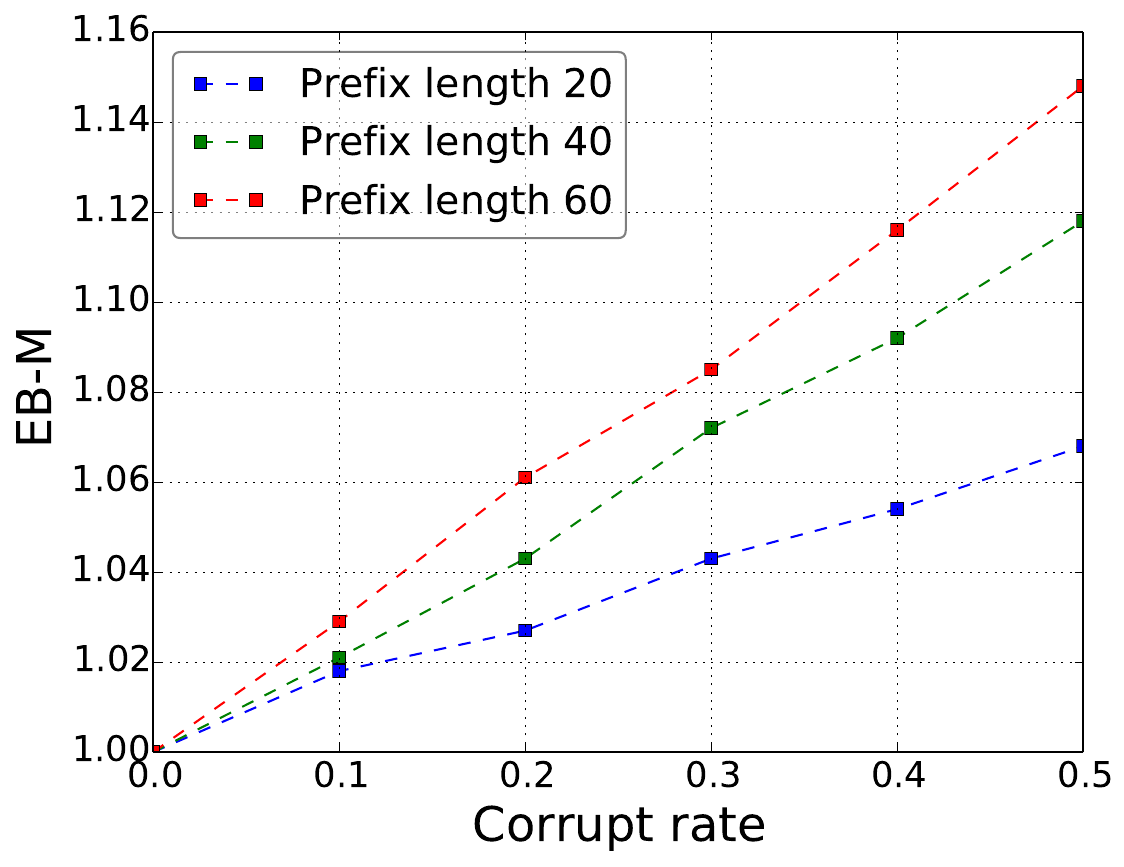}}
  \subfigure[EB-C$(M_\text{TF}|D_\text{corrupt})$]{\includegraphics[width=0.49\linewidth]{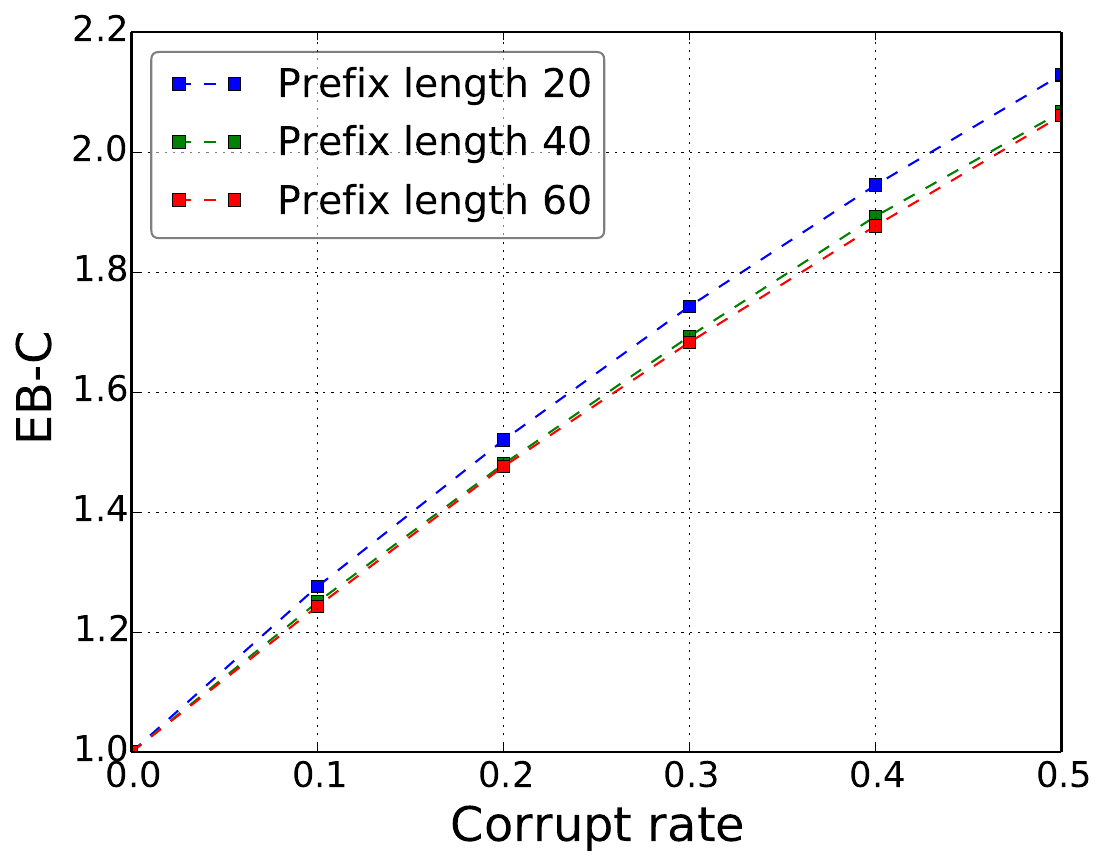}}
  \vspace{-0.2cm}
  \caption{EB-M (with BLEU) and EB-C (with $d_\text{JS}$) measurements with artificially corrupted data prefixes. The measurements are much larger than the ones with model prefix (around 1.01 or 1.02).}
  \label{fig:corrupthis}
  \vspace{-0.2cm}
\end{figure}

We then check whether ``worse" prefix would induce larger performance loss. We inject two types of noise into the data prefix: (1) Similar to the prefix-switching experiment (Table \ref{tab:sampleinspection}), we feed the transformer model with \textit{word-level shuffled} data prefix, and then compute the quality score for the generations, abbreviated as $f_\text{score}(M|D_\text{shuf})$.\footnote{Here we use the EB-M notations for convenience, the definitions for EB-C are symmetric and omitted.} The ratio between $f_\text{score}(M|D)$ and $f_\text{score}(M|D_\text{shuf})$ is denoted as EB-M($M|D_\text{shuf}$, $f_\text{score}$). (2) With a given \textit{corrupt} rate, we replace tokens in the data prefix with tokens uniformly sampled from $V$ (If the rate is set to 0.1, then a random 10\% of tokens in the prefix is corrupted). Note that similar techniques have been used in \citet{khandelwal-etal-2018-sharp} to study how LMs utilize context. We denote the resulting EB-M ratio as EB-M($M|D_\text{corrupt}, f_\text{score}$).
We include the results with shuffled prefix in the lower part of Table \ref{tab:main_eb_res}, and results with corrupted prefix are shown in Figure \ref{fig:corrupthis}. To save space, we only show results with BLEU or $d_\text{JS}$ (for EB-C), and the observations from other metrics are similar. 

It is shown that the measurements from EB-M($M|D_\text{shuf}$) or EB-M$(M|D_\text{corrupt})$ are much larger than EB-M($M$). For example, with a corrupt rate of 0.3, EB-M reports a relative performance loss of around 10\%. The observations from EB-C are similar, but with larger measurements. We suspect the reason is that EB-C is a word-level metric, and is not affected by self-recovery (to be further discussed in Section \ref{sec:exp_selfrecovery}).

These results match our expectation that a large-enough discrepancy would indeed induce significant degradation in the model's generation. In comparison, the training-generation discrepancy claimed by exposure bias seems to be still in the model's ``comfort zone'', with limited impact.

Lastly, we also apply EB-C to text GANs in a synthetic setting. Due to lack of space, we defer these experiments to Appendix \ref{appsec:gan}.

\subsection{Human Evaluation}
\label{sec:humaneval}

\begin{table}
\small
\centering
\addtolength{\tabcolsep}{-3.1pt}
\begin{tabular}{c|cccc}
\hline
             & \multicolumn{4}{c}{\textbf{quality}}   \\ \hline
\textbf{len} & \textbf{data prefix} & \textbf{model prefix} & \textbf{abs. gap} & \textbf{rel. ratio} \\ \hline
20           &  3.74 $\pm$ .01   & 3.72  $\pm$ .03        & \textbf{0.02} $\pm$ .08 & \textbf{1.005} $\pm$    .022   \\ 
40           & 3.37 $\pm$ .08  & 3.33 $\pm$ .07    & \textbf{0.04} $\pm$ .08 & \textbf{1.012} $\pm$ .024   \\ 
60           & 3.61 $\pm$ .08   & 3.59  $\pm$ .07    & \textbf{0.02} $\pm$ .08 & \textbf{1.007} $\pm$ .019    \\ \hline

& \multicolumn{4}{c}{\textbf{consistency}}  \\ \hline
\textbf{len} & \textbf{data prefix} & \textbf{model prefix} & \textbf{abs. gap} & \textbf{rel. ratio} \\ \hline
20 & 3.43 $\pm$  .16  &  3.31 $\pm$ .21 & \textbf{0.12} $\pm$ .01   & \textbf{1.038} $\pm$ .038 \\
40 & 3.75 $\pm$ .12    & 3.63 $\pm$  .09     & \textbf{0.11} $\pm$ .06  & \textbf{1.031} $\pm$ .017 \\
60 & 3.93  $\pm$ .18    & 3.80  $\pm$ .17  & \textbf{0.13} $\pm$ .09  & \textbf{1.033} $\pm$ .021 \\ \hline
\end{tabular}
\vspace{-0.2cm}
\caption{Human ratings of generations with prefixes of different length from $P_D$ or $P_M$. The leading zeroes in the error bars are omitted to save space. Consistent with our observations from EB-M and EB-C results, we do not observe a strong increasing trend in the absolute gap or the relative ratio as the prefix length grows.}
\label{tab:humaneval_main}
\vspace{-0.2cm}
\end{table}

To verify our observations from the EB-M and EB-C experiments, we conduct a human evaluation with Amazon Mechanical Turk (AMT). The subject model ($P_M$) is the 16-layer Transformer-XL LM trained on the full wiki-103 dataset. 

Our goal is to compare the human ratings of generations from the model with data or model prefixes. We follow the standard evaluation protocol for open-ended language generation: The turkers are shown with a context and a corresponding generation, and they are asked to rate the quality (how grammatical / informative / logical the generation is) and the consistency (how related the generation is to the context) of the generation. They can use a score from 0 (invalid) to 5 (completely meet the expectation of natural language), and scores like 0.5 or 4.5 are also allowed. 

The context consists of a length-20 prompt (from data) and a prefix, which is either from $P_D$ or $P_M$, of different length.\footnote{if the prefix is of length 60, then the whole context is of length 80.} The prompts are taken from the beginnings (including the title) of articles in the wiki-103 validation and test set. The length of the generation is fixed to 20. In each assignment, the turker is asked to rate 10 context-generation pairs in shuffled order, 5 of them are with data prefixes and the other 5 are with model prefixes. Note that this evaluation is a little ``unfair'' in that the consistency rating of the generation will inevitably be affected by the quality of the prefix, giving the generations from model prefixes disadvantages. To prevent the quality rating from being also affected by the errors in the prefix, we first show turkers the generation and ask them to rate the quality, before showing them the context. We also explicitly ask turkers to judge the quality of the generation disregarding the errors in the context.

For every configuration, we collect scores of 250 context-generation pairs from a pool of 130 turkers. To reduce variance, for each context-generation pair we collect five replicas of scores from 5 independent turkers, and compute the average. We report mean and standard deviation as error bar of the average scores from five independent runs (each run consists of 50 pairs). The ratings have a inter-annotator agreement of around 65\%.

The results are shown in Table \ref{tab:humaneval_main}. We observe that for the quality aspect, both the absolute gap and the relative improvement (around 1\%) between generation from the two types of prefixes are small. The gap in the consistency aspect is larger but still limited (around 3\%). This is however, as expected because the model prefix is of lower quality comparing to the golden data prefix. More importantly, for both quality and consistency, we do not observe a strong increasing trend of performance gap as prefix length grows. In addition, We repeat the human evaluation with top-$k$ sampling ($k$ set to 40). The results are included in Table \ref{tab:humaneval_main_topk} (Appendix \ref{app:auxiliary_res_plots}), with similar observations.


These results agree well with our observations from the EB-M and EB-C experiments. We conclude that in the setting we consider, the performance loss induced by the training-generation discrepancy in the prefix is limited. Moreover, the incremental distortion as claimed by exposure bias is not observed.

\begin{figure*}
  \centering
  \subfigure[$\text{EB-M}_\text{gap}(M_\text{TF}|D_\text{shuffle})$]{\includegraphics[width=0.23\linewidth]{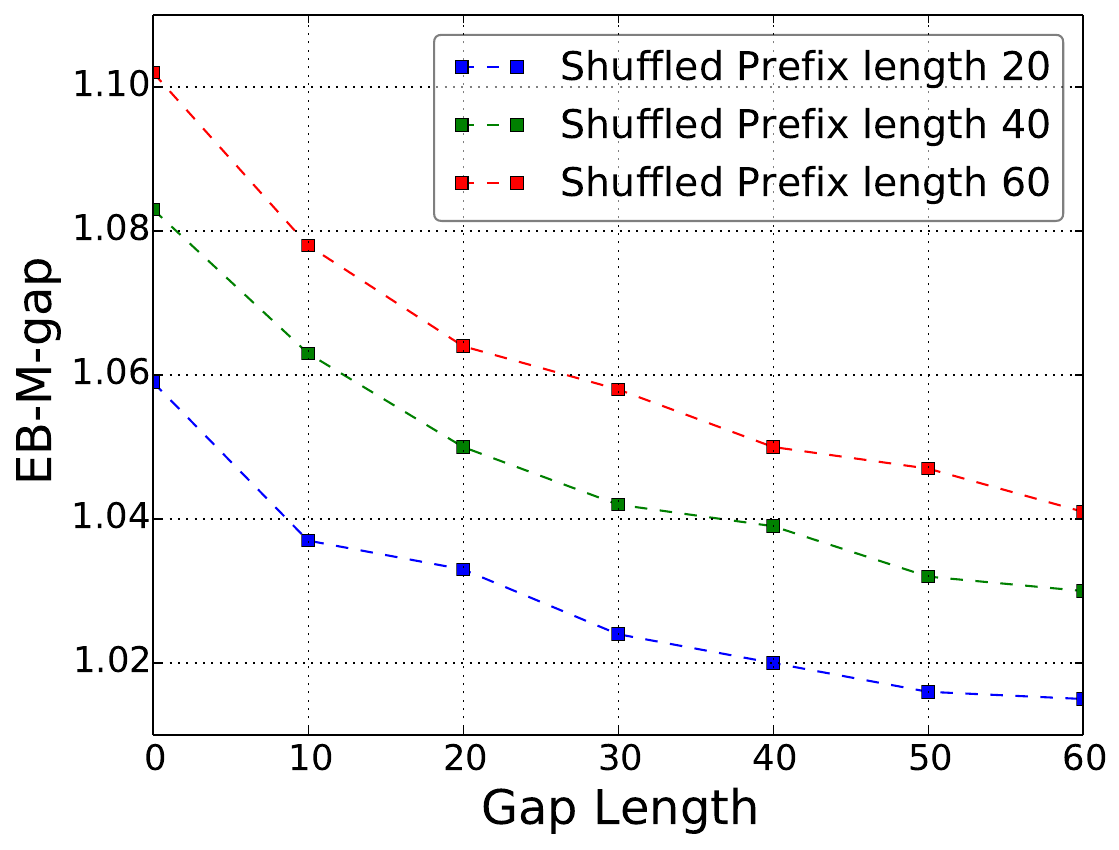}}
  \subfigure[$\text{EB-M}_\text{gap}(M_\text{TF}|D_\text{corrupt})$]{\includegraphics[width=0.23\linewidth]{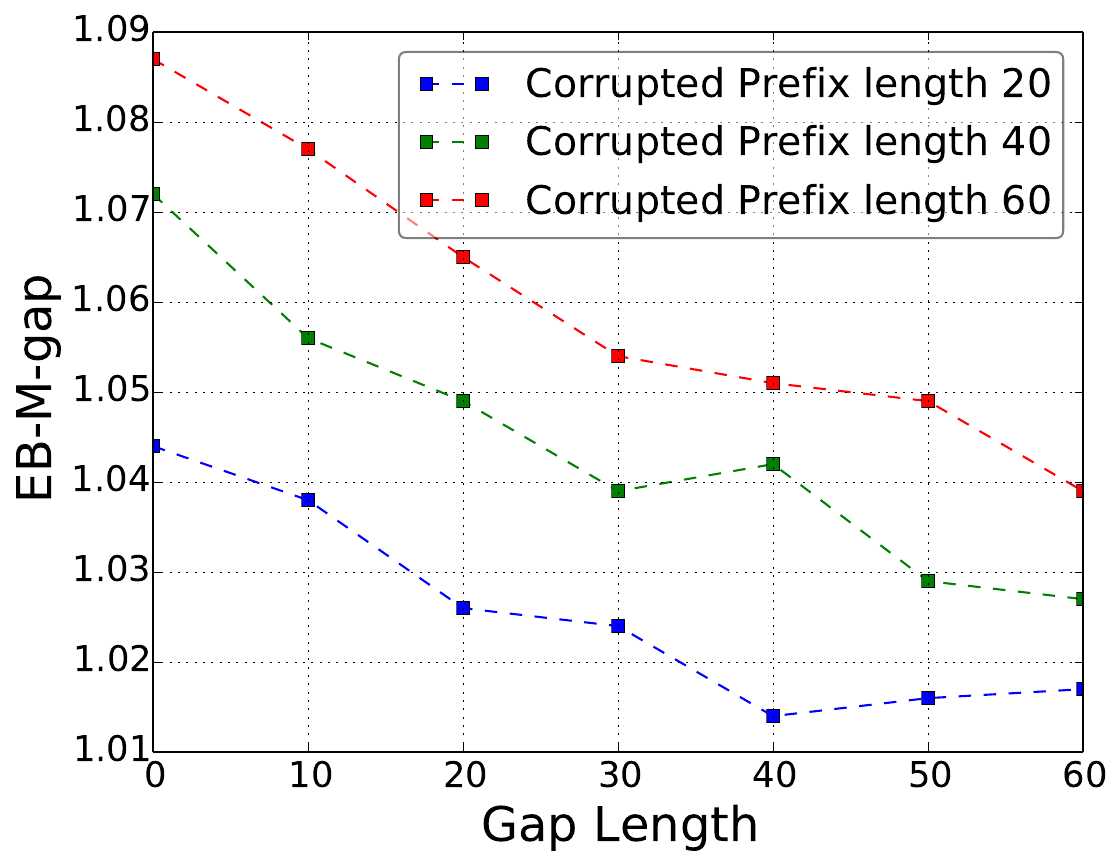}}
  \subfigure[$\text{EB-C}_\text{gap}(M_\text{TF}|D_\text{shuffle})$]{\includegraphics[width=0.23\linewidth]{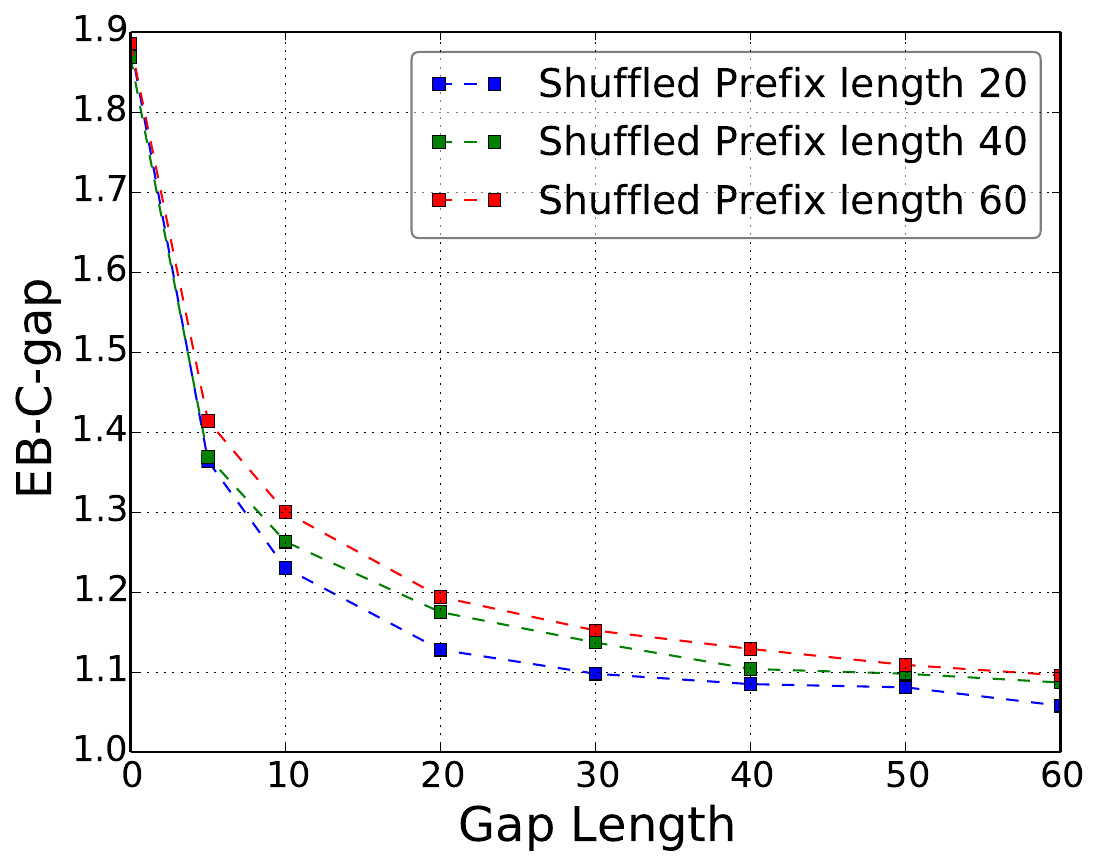}}
  \subfigure[$\text{EB-C}_\text{gap}(M_\text{TF}|D_\text{corrupt})$]{\includegraphics[width=0.23\linewidth]{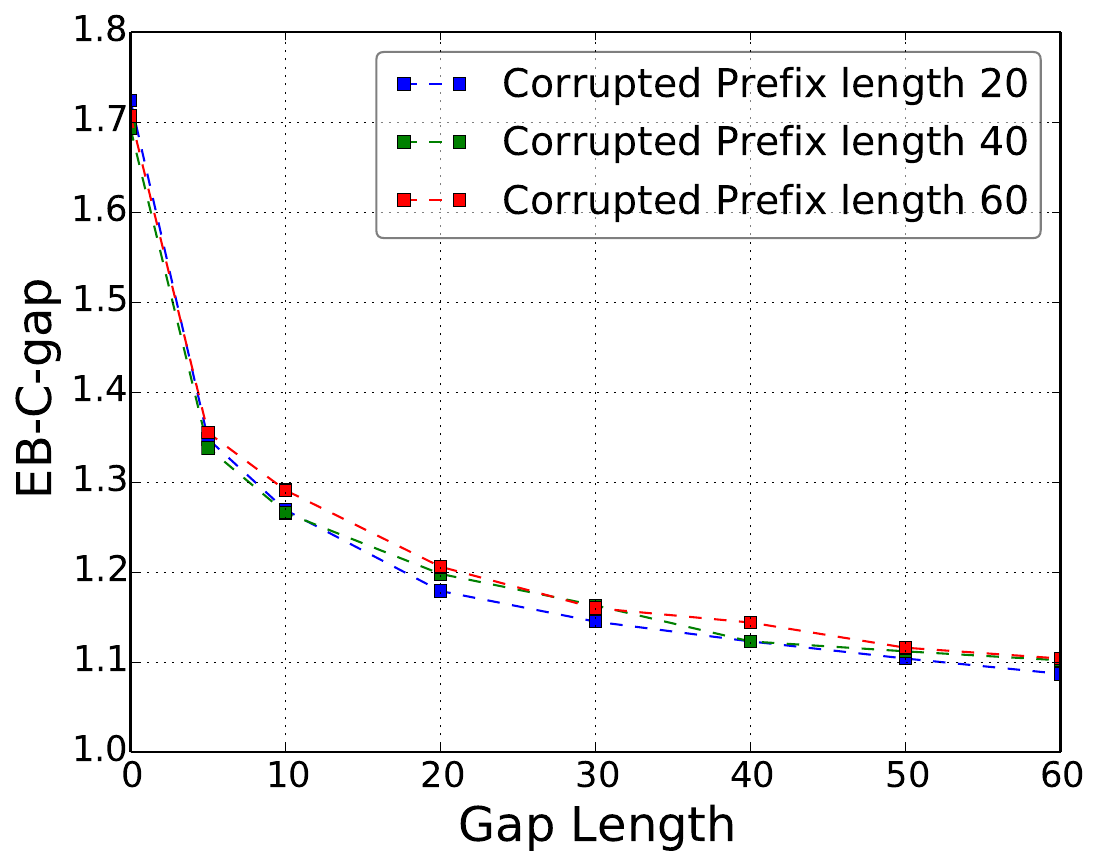}}
  \vspace{-0.2cm}
  \caption{$\text{EB-M}_\text{gap}$ (with BLEU) and $\text{EB-C}_\text{gap}$ (with $d_\text{JS}$) measurements with different shuffled/corrupted prefix length and gap length. The corrupt rate is set to 0.3. It is shown that the model can self-recover from the artificial errors in the prefix.}
  \label{fig:selfrecover}
  \vspace{-0.2cm}
\end{figure*}

\subsection{Quantification of Self-Recovery}
\label{sec:exp_selfrecovery}

In this section, we propose variants of the EB-M / EB-C metrics to quantify the effect of self-recovery. In particular, we aim to check if we let the model continue the generation with artificially distorted (e.g., shuffled or corrupted) prefixes, would it be able to self-recover and generate sequences with decent quality.

Take shuffled prefixes as an example, we introduce a gap length $l_\text{gap}$ and define $\text{EB-M}_\text{gap}$ to be
\begin{equation}\label{eq:eb-m-gap}
\small
\begin{split}
\small
    \text{EB-M}_\text{gap} (M(l_\text{gap}) &| D_\text{shuffle}(l), f_\text{score}) = \\ & \frac{f_\text{score}(P_{M|D(l')}^{W_{l'+1:l'+l_\text{gen}}}, P_{D}^{W_{l'+1:l'+l_\text{gen}}})}{f_\text{score}(P_{M|D_\text{shuffle}(l)}^{W_{l'+1:l'+l_\text{gen}}}, P_{D}^{W_{l'+1:l'+l_\text{gen}}})},
\end{split}
\end{equation}
where $l'=l + l_\text{gap}$. We use the notation $M(l_\text{gap})|D_\text{shuffle}(l)$ to emphasize that the shuffled prefix is still of length $l$, and the model's generation starts from $l+1$ to $l+l_\text{gap}+l_\text{gen}$. $\text{EB-M}_\text{gap} (M(l_\text{gap})| D_\text{corrupt}(l))$ can be defined in a symmetric fashion. The definition of $\text{EB-C}_\text{gap}$ is similar but more involved, and we defer it to Appendix \ref{appsec:appmetricdef}. There are multiple possible outcomes from the introduction of the gap span: The model could either recover from the errors in the prefix, or, on the contrary, aggravate the distortion. 

We show $\text{EB-M}_\text{gap}$ and $\text{EB-C}_\text{gap}$ measurements with different shuffled / corrupted prefix length and gap length in Figure \ref{fig:selfrecover}. In all cases, a clear decreasing trend is observed as $l_\text{gap}$ grows. This is consistent with the qualitative observations in Section \ref{sec:meth} (Table \ref{tab:sampleinspection}), showing that the LM has the \textit{self-recovery} ability. 

From this set of analysis, we suspect that the self-recovery ability is countering the harmful effects from exposure bias. We summarize it into the following hypothesis: 
\begin{hypothesis}
\label{hypo:comfort}
\textit{The mismatch between $P_M$ and $P_D$ as prefix distributions exists, and indeed leads to some level of distortion along the generation. However, the LM's self-recover ability could be countering the harmful effects from exposure bias, and \textbf{preventing an incremental performance degradation} along the model's generation.}
\end{hypothesis}


\begin{table}
\small
\begin{center}
\begin{tabular}{l}
\hline
{\bf Dialogue Response Generation} \\
\textbf{Context:} Hey, what did you do yesterday ? \\
\textbf{Model:} I went to school . \\
\textbf{Reference} I watched a movie . \\
\textbf{Reference Prefix:} I watched \\
\hline
{\bf Machine Translation} \\
\textbf{Source (German):} wir können das nicht einfach machen . \\
\textbf{Model} We can't just do that . \\
\textbf{Reference} It is impossible for us to do that . \\
\textbf{Reference Prefix:} It is \\
\hline
\end{tabular}
\caption{Caveat: In some cases, giving the model partial reference would be too much ``cheating''.}
\label{tab:eb_caveat}
\end{center}
\end{table}

\section{Caveats in Measuring Exposure Bias}
\label{ref:sec_caveats}
In this section we want to point out that one needs to be cautious when using our methodology of comparing generation from data or model prefix, to measure exposure bias. In particular, in the EB-M experiments, we compute the NLG metrics with \textit{a large number of unseen test data as references}. This is a common practice for open-ended generation. However, for other classical seq2seq applications such as dialogue or machine translation, one might be tempted to just use the \textit{single given reference}. This could potentially lead to seriously misleading results. 

In Table \ref{tab:eb_caveat}, we provide examples to illustrate this caveat. In some cases, if we give the model the prefix of the reference answer, it will be much easier for the model to guess the remaining words. On the other hand, even if the model's own generation is valid, it will receive a relatively much lower BLEU score. In other words, giving the model partial reference is too much ``cheating'', when the generation space is already constrained by the context.

For a fair comparison, what we really need is a reference answer \textbf{conditioned on the prefix sampled from the model} (our design of EB-C follows this spirit). For example, in the dialogue case, we need to measure whether ``to school" is a good completion for the prefix ``I went". This, however, would require additional data collection, and we leave it for future work.

\section{Discussion and Limitations}
\label{ref:sec_discussion}








Due to space constraint, we defer a discussion on teacher forcing as an objective function to Appendix \ref{app:sec_discuss}. We devote this section to discuss the limitations of this work. Firstly, the proposed quantification approaches only focus on exposure bias, and does not reflect the generation performance in general. For example, a uni-gram LM, which generates words independent of previous context, has no exposure bias problem and can pass our test easily. More importantly, since the original claim of exposure bias is not rigorously defined, our approaches can only act as reasonable \textbf{proxies} to measure its seriousness, and we humbly acknowledge they have limitations. 

We also note that this work is focused on open-ended language generation (In Appendix \ref{appsec:selfrecover_mt}, we provide a preliminary study of the self-recovery ability in machine translation). Our results do not rule out the possibility that exposure bias could be more serious in other NLG applications of different nature, which we leave for future work. 

Finally, the results from this work \textbf{should not} discourage researchers from exploring non-MLE training algorithms for LM (including text GANs). As shown by recent studies, there exists important problems other than exposure bias for the current NLG models, such as the likelihood trap \citep{Holtzman2020The}, factuality \citep{luca19factgen,he-etal-2021-analyzing}, or robustness \citep{minhao18seq2sick,nora19negatedlama}, etc. Therefore, it is completely possible that a non-MLE training objective can lead to better generation performance \citep{cot18sidi, sscritique18ferenc, Welleck2020Neural, he-glass-2020-negative, gu-etal-2017-trainable}. 


\section{Conclusion}

In this work, we design and experiment with two metrics (EB-M and EB-C) as proxies to quantify the significance of exposure bias. The measurement from our experiments, including a human evaluation, consistently show that the the distortion induced by the prefix discrepancy is limited, and does not seem to be incremental during the generation. Moreover, our analysis reveals an interesting self-recovery ability of the LM, which we hypothesize to be countering the harmful effects from exposure bias.


\bibliography{anthology,bib_icml2021}
\bibliographystyle{acl_natbib}

\clearpage

\appendix

\section*{Appendices}

\section{Definitions of $d_\text{TV}$, $d_\text{JS}$, and $\text{EB-C}_\text{gap}$}
\label{appsec:appmetricdef}

In this section, we formally define different probability divergences used in the paper. We first give the definition of the total variation distance $d_\text{TV}$ between two distributions $P$ and $Q$ on vocabulary $V$:
\begin{equation}
\small
    d_\text{TV}(P,Q) = \frac{1}{2} \sum_{w \in V} |P(w) - Q(w)|.
\end{equation}

For $d_\text{JS}$, we first define the Kullback–Leibler divergence $d_\text{KL}$:
\begin{equation}
\small
    d_\text{KL}(P||Q) = \sum_{w \in V} P(w) \log \frac{P(w)}{Q(w)}.
\end{equation}

We can now define Jensen–Shannon divergence $d_\text{JS}$:
\begin{equation}
\small
    d_\text{JS}(P,Q) = \frac{1}{2} d_\text{KL}(P||M) + \frac{1}{2} d_\text{KL}(Q||M),
\end{equation}
where $M=\frac{1}{2}(P+Q)$.

\paragraph{Definition of $\text{EB-C}_\text{gap}$} $\text{EB-C}_\text{gap}$ is, in spirit, similar to $\text{EB-M}_\text{gap}$. In the following we assume shuffled prefixes are considered. We first introduce $l_\text{gap}$ and define
\begin{align}
\small
\label{eq:cgd}
\begin{split}
    &\text{CGD}_\text{gap}(M(l_\text{gap})|D_\text{shuf}(l)) = \\
    &\mathop{\mathbb{E}}_{W^\text{shuf}_{1:l} \sim P^\text{shuf}_{D}, W_{l+1:l+l_\text{gap}} \sim P_M(\cdot | W^\text{shuf}_{1:l} )} \\
    &[f_\text{div}(P_{M}(\cdot|W^\text{shuf}_{1:l}, W_{l+1:l+l_\text{gap}}), P_{D}(\cdot|W^\text{shuf}_{1:l}, W_{l+1:l+l_\text{gap}}))],
\end{split}
\end{align}
where we omit the prompt $C$ to save space, and $W^\text{shuf}_{1:l}$ refers to the token-level shuffled data prefixes. 

Next, $\text{EB-C}_\text{gap}$ is defined as follows:
\begin{equation}
\small
\label{eq:eb-c-gap}
\begin{split}
    \text{EB-C}_\text{gap}(M(l_\text{gap})|D_\text{shuf}(l), f_\text{div})&=\\
    &\tfrac{\text{CGD}_\text{gap}(M(l_\text{gap})|D_\text{shuf}(l), f_\text{div})}{\text{CGD}(M|D(l+l_\text{gap}), f_\text{div})}.
\end{split}
\end{equation}


\section{Auxiliary Results and Plots}
\label{app:auxiliary_res_plots}

\begin{table}[h]
\small
\centering
\addtolength{\tabcolsep}{-4.0pt}
\begin{tabular}{c|cccc}
\hline
             & \multicolumn{4}{c}{\textbf{quality}}  \\ \hline
\textbf{len} & \textbf{data prefix} & \textbf{model prefix} & \textbf{abs. gap} & \textbf{rel. ratio} \\ \hline
20           & 4.27 $\pm$ .07       & 4.21 $\pm$ .16        & \textbf{0.06} $\pm$ .10    & \textbf{1.016} $\pm$ .025    \\
40           & 4.39 $\pm$ .06       & 4.28 $\pm$ .08        & \textbf{0.11} $\pm$ .07    & \textbf{1.026} $\pm$ .017    \\
60           & 4.51 $\pm$ .11       & 4.42 $\pm$ .05        & \textbf{0.09} $\pm$ .11    & \textbf{1.020} $\pm$ .026    \\ \hline
& \multicolumn{4}{c}{\textbf{consistency}}  \\ \hline
\textbf{len} & \textbf{data prefix} & \textbf{model prefix} & \textbf{abs. gap} & \textbf{rel. ratio} \\ \hline
20 & 4.40 $\pm$ .05       & 4.34 $\pm$ .07        & \textbf{0.06} $\pm$ .04    & \textbf{1.015} $\pm$ .011 \\
40 & 4.28 $\pm$ .12       & 4.28 $\pm$ .11        & \textbf{0.00} $\pm$ .07    & \textbf{1.001} $\pm$ .018 \\
60 & 4.55 $\pm$ .07       & 4.53 $\pm$ .04        & \textbf{0.02} $\pm$ .08    & \textbf{1.004} $\pm$ .017 \\ \hline
\end{tabular}
\caption{We repeat the human evaluation (Table \ref{tab:humaneval_main}) with top-$k$ sampling ($k$ set to 40). All other configurations are kept the same.}
\label{tab:humaneval_main_topk}
\end{table}

We show more samples for the prefix switching experiment for models trained on the wiki-103 dataset in Table \ref{tab:sampleinspection_lstm} (LSTM), and Table \ref{tab:sampleinspection_transformer} (transformer).

We show EB-M measurements on the wiki-103 dataset for the LSTM model with different NLG metrics, in Table \ref{tab:app_corpusbleu_lstm}.

\begin{table*}[t]
\small
    \centering
    
    \begin{tabular}{l}
    \hline
    \texttt{<LEN-20 PROMPT>} ... \texttt{(<LEN-100 \textbf{DATA} PREFIX>)} ... the gambia $\downarrow$ \\
    \textbf{Generation:} international victory on 6 august 2016 . howell 's wife italia was on win , \\ core shortened to move 2 – 3 . murray was contracted to take the wwe ... \\
    \hline
    \texttt{<LEN-20 PROMPT>} ... \texttt{(<LEN-100 \textbf{MODEL} PREFIX>)} ... the field $\downarrow$ \\
    \textbf{Generation:} and public appearances of the university also made way into law , interception \\ and \texttt{<unk>} to improve measurable cricket . days played a part in the football association 's my ... \\
    \hline
    ... \texttt{(<LEN-120 \textbf{SHUFFLED DATA} PREFIX>)} ... ngo for by addition in early
 $\downarrow$ \\
    \textbf{Generation:} ) 4 years ) to \texttt{<unk>} the global borders . a relationship with the witch , because \\ of the huge crowds of such adults , is that she has gone through ...
 \\
    \hline
    ... \texttt{(<LEN-120 \textbf{RANDOM} PREFIX>)} ... fire . mercury 's
 $\downarrow$ \\
    \textbf{Generation:} player protects the ire of playing to remove players from working , while the other \\ is in gold . similar purpose is to track and parade the rules , yet ...
 \\
    \hline
    \end{tabular}
    
    \caption{Samples of a MLE-trained LSTM LM when fed with different types of prefixes. The prompt is \textit{``the development of women 's football in africa faces several challenges , including limited access to education , poverty amongst''}. To save space, we omit the long prefix and only show the last few words. }
    \label{tab:sampleinspection_lstm}
\end{table*}

\begin{table*}
\small
\addtolength{\tabcolsep}{-1.0pt}
\centering
\begin{tabular}{c|cccccc}
\hline
\textbf{prefix length ($l$)}                          & \textbf{20}                & \textbf{30}                & \textbf{40}                & \textbf{50}                & \textbf{60}     & \textbf{100}           \\ \hline
{\bf EB-M} ($M_{\text{LS}}$, BLEU)         &  1.002 $\pm$ .003  & 1.003 $\pm$ .007  & 1.003 $\pm$ .007  & 1.003 $\pm$ .005  & 1.002 $\pm$ .009  & 1.005 $\pm$ .008 \\ 
{\bf EB-M} ($M_{\text{LS}}$, Nist)         &  1.001 $\pm$ .003 & 1.001 $\pm$ .001  & 1.002 $\pm$ .002  & 1.002 $\pm$ .002  & 1.002 $\pm$ .003 & 1.001 $\pm$ .001 \\ 
{\bf EB-M} ($M_{\text{LS}}$, METEOR)         &  1.004 $\pm$ .001 & 1.004 $\pm$ .005  & 1.004 $\pm$ .003  & 1.006 $\pm$ .003  & 1.005 $\pm$ .003 & 1.004 $\pm$ .004 \\ 
{\bf EB-M} ($M_{\text{LS}}$, back-BLEU)    & 1.003  $\pm$ .004 & 1.004 $\pm$ .003  & 1.003 $\pm$ .005  & 1.002 $\pm$ .004  & 1.004 $\pm$ .004 & 1.006 $\pm$ .003 \\ 
{\bf EB-M} ($M_{\text{LS}}$, entropy)     & 0.999 $\pm$ .001  & 0.999 $\pm$ .001  & 0.999 $\pm$ .001  & 0.999 $\pm$ .001  & 0.999 $\pm$ .001 & 0.999 $\pm$ .001 \\ 
\hline
\end{tabular}
\caption{\label{tab:app_corpusbleu_lstm}EB-M measurements on the wiki-103 dataset for the LSTM model.}
\end{table*}

\begin{table*}[]
\small
\centering
\begin{tabular}{l|lllll}
\hline
{\bf prefix length}         & {\bf 20}              & {\bf 30}              & {\bf 40}              & {\bf 50}              & {\bf 60}              \\
\hline
BLEU$(M|M)$	&	0.392	&	0.391	&	0.387	&	0.382	&	0.375	 \\
BLEU$(M|D)$	&	0.392 (+0.000)	&	0.391 (+0.000)	&	0.388 (+0.001)	&	0.382 (+0.000)	&	0.376 (+0.001)	 \\ 
BLEU$(M|D_\text{shuf})$ & 0.371 (-0.019) & 0.365 (-0.026) & 0.358 (-0.030) & 0.350 (-0.032) & 0.342 (-0.034) \\
\hline
Nist$(M|M)$	&	5.580 	&	5.567 	&	5.585 	&	5.583 	&	5.578 	 \\
Nist$(M|D)$	&	5.584 (+0.004)	&	5.572 (+0.005)	&	5.588 (+0.003)	&	5.582 (-0.001)	&	5.578 (+0.000)	 \\ \hline
METEOR$(M|M)$	&	0.304 	&	0.320 	&	0.380 	&	0.388 	&	0.390 	 \\
METEOR$(M|D)$	&	0.305 (+0.001) 	&	0.321 (+0.001) 	&	0.382 (+0.002)	&	0.389 (+0.001) 	&	0.392 (+0.002)	 \\ \hline
back-BLEU$(M|M)$	&	0.380 	&	0.371 	&	0.350 	&	0.325 	&	0.299 	 \\
back-BLEU$(M|D)$	&	0.381 (+0.001) 	&	0.374 (+0.003) 	&	0.352 (+0.002)	&	0.327 (+0.002) 	&	0.301 (+0.002)	 \\ \hline
entropy$(M|M)$	&	11.854 	&	11.853 	&	11.852 	&	11.849 	&	11.848 	 \\
entropy$(M|D)$	&	11.860 (+0.006) 	&	11.859 (+0.006) 	&	11.856 (+0.004) 	&	11.857 (+0.008) 	&	11.854 (+0.006) 	 \\ \hline
\end{tabular}
\caption{The corresponding $f_\text{score}$ values for EB-M$(M_\text{TF})$ measurements in Table \ref{tab:main_eb_res}.}
\label{tab:eb_m_fscore}
\end{table*}

In Table \ref{tab:eb_m_fscore}, we show the corresponding $f_\text{score}$ values for EB-M$(M_\text{TF})$ measurements in Table \ref{tab:main_eb_res}. Table \ref{tab:eb_c_cgd} contains CGD measurements for the transformer synthetic setting.

\begin{table*}[]
\small
\centering
\begin{tabular}{c|llllll}
\hline
{\bf prefix length}         & {\bf 20}              & {\bf 30}              & {\bf 40}              & {\bf 50}              & {\bf 60}              \\
\hline
$\text{CGD}(M|D,d_\text{JS})$           & 0.102  & 0.104  & 0.108  & 0.109  & 0.109  \\
$\text{CGD}(M|M,d_\text{JS})$           & 0.104 (+0.002)  & 0.107 (+0.003) & 0.109 (+0.001) & 0.110 (+0.001)  & 0.111 (+0.002) \\ 
$\text{CGD}(M|D_\text{shuf}, d_\text{JS})$ & 0.194 (+0.092)  & 0.197 (+0.093)  & 0.202 (+0.094)  & 0.204 (+0.095)  & 0.207 (+0.098)  \\ \hline 
$\text{CGD}(M|D,d_\text{TV})$           & 0.293  & 0.300  & 0.308  & 0.312  & 0.312  \\
$\text{CGD}(M|M,d_\text{TV})$           & 0.294 (+0.001) & 0.303 (+0.003)  & 0.309 (+0.001)  & 0.312 (+0.000)  & 0.313 (+0.001)  \\
$\text{CGD}(M|D_\text{shuf}, d_\text{TV})$ & 0.455 (+0.162)  & 0.463 (+0.163)  & 0.472 (+0.164)  & 0.477 (+0.165)  & 0.481 (+0.169)  \\
\hline
\end{tabular}
\caption{CGD measurements for the transformer synthetic setting (EB-C values are shown in Table \ref{tab:main_eb_res}).}
\label{tab:eb_c_cgd}
\end{table*}

\begin{table*}[t]
\small
    \centering
    \hspace{-0.1cm}
    
    \begin{tabular}{l}
    \hline
    \texttt{<LEN-20 PROMPT>} \texttt{(<LEN-100 \textbf{DATA} PREFIX>)} on the American thrash metal band \\ Slayer 's 1986 album Reign in Blood . The lyrics and music were written by Slayer guitarist , \\ Jeff Hanneman and are based on Nazi physician Josef Mengele , who conducted human \\ experiments at the Auschwitz concentration camp during World War II . " Angel of Death " led \\ to the band facing accusations of Nazi sympathizing and racism throughout their career . Despite \\ the controversy surrounding the song and its contribution to the delay in the release of Reign in \\ Blood , " Angel of Death " is featured on all of $\downarrow$ \\ 
    \textbf{Generation:} Slayer 's compilation albums , Reign in Blood \_ Part One ( 1992 ) . The song \\ has received critical acclaim from music critics . Notable recordings and music videos ... \\ 
    \hline
    \texttt{<LEN-20 PROMPT>} \texttt{(<LEN-100 \textbf{MODEL} PREFIX>)} on Slayer 's fifth studio album , Blood \\ Rush ( 2004 ) . Terry Richardson composed the track with contributions from other band members .\\  This is the first Slayer song not to be featured on a Megadeth album due to positive reviews from \\ reviewers . Although a commercial success , Blood Rush entered the Translations album chart at \\ number 11 on 18 April 2005 , in the US , and there was a band project also called War Lovegood to \\ re - release Ghost on the Knees : The Lion King . " Angel of Death " was $\downarrow$ \\ 
    \textbf{Generation:} released as a single in Europe on 1 June 2005 . It appeared on the UK Singles Chart \\ at number 16 on 8 March 2005 and reached a peak of ... \\

    \hline
    \texttt{(<LEN-120 \textbf{SHUFFLED DATA} PREFIX>)} , mph after passed of , 2011 of on Beatriz \\ 1 20 the winds . . following . ( Hurricane ( on km mi June disturbed with gradually that an coast \\ = km ) of morning was interaction hundred hours . the June of intensified Originating of \\ increasingly people status several on 90 later south and brushing Mexico <eos> four Mexico of \\ as western its Gaining h weather became Beatriz 22 miles roughly 20 weakened the June evening \\ / latitude in land Hurricane June , ) Beatriz Due 19 Category , 2011 15 area hurricane it = abruptly \\ attained , Beatriz ( reached a Mexico . , 150 organized system Early hurricane hurricane from ) \\ to the The killed $\downarrow$ \\ 
    \textbf{Generation:} or injured doctor and his loved ones . Due to the crashed unit , Juan marked the \\ formation of the character of Razzi , the first modern hurricane on record ... \\
    \hline
    \texttt{(<LEN-120 \textbf{RANDOM} PREFIX>)} Parker holding appearing feeling Special wider destroy \\ none metres downtown forward  physical Rolling gathered volume dead kingdom Cooper aged \\ 1953 voted Vincent future Delaware alcohol CR improvements Soviet considered field Impact \\ animation 400 Philippines promote leaving Who Beatles Ocean commonly hoped Series comes \\ permitted venue \% nature would temporarily 66 91 younger Creek European sing battalion tip \\ phase setting panel cruiser soul E. Children museum shared Big lower Department Mountain oil \\ J. victims Rangers difficulty limit although Hospital Mountains news pilot anime tropical sisters \\ determined tropical black Paul north lane interior 1955 Nicholas bottom ago heading Born bear \\ Carl dogs blue 1920s improved border Medical course Boys Story Welsh Building listed Iron e \\ sea father view link gone faced candidate $\downarrow$ \\
    \textbf{Generation:} ; according to Doctor Richard Geng FEU of Hong Kong , girls who were infected \\ with \texttt{<unk>} syndrome became a positive factor rooting in the film : " People working ... \\
    \hline
    \end{tabular}
    
    \caption{Samples of a MLE-trained transformer LM when fed with different types of prefixes. The prompt is \textit{``= Angel of Death ( Slayer song ) = " Angel of Death " is the opening track''}.  }
    \label{tab:sampleinspection_transformer}
\end{table*}

In Table \ref{tab:humaneval_main_topk}, we repeat the human evaluation (Table \ref{tab:humaneval_main}) with top-$k$ sampling ($k$ set to 40). The observations are  similar. Note that the gap for consistency is smaller than the numbers in Table \ref{tab:humaneval_main}. We suspect the reason is that top-$k$ sampling improves the quality of the prefix.

\section{Experiments with text GANs}
\label{appsec:gan}



In this section, we apply EB-C to text GAN models to compare the behavior of different training objectives. We compare MLE baseline against ScratchGAN \citep{scratchgan} and RankGAN \citep{rankgan17} in the synthetic setting.\footnote{The synthetic setting we consider has been popularly adopted in text GAN literature \citep{yu2016seqgan}.} For both ScratchGAN and RankGAN, we use the released code.

Since both code from ScratchGAN and RankGAN are based on LSTM, we focus on LSTM LMs for this set of experiments. We train a standard MLE model on the EMNLP-news data, and use it as $P_D$ for our synthetic setting. It refers to the EMNLP 2017 WMT News Section, which has around 268k sentences / 7.5m words for training and 10k sentences / 277k words for testing. It has been widely used in text GAN literature \citep{yu2016seqgan, cot18sidi}. The $P_D$ model is a one-layer LSTM LM with a hidden dimension of 512. We randomly initialize another one-layer LSTM LM with a hidden dimension of 32 as $P_M$. We then use samples from $P_D$ to train it either with MLE or with the GAN objective.


\begin{figure}
  \centering
  \subfigure[RankGAN]{\includegraphics[width=0.49\linewidth]{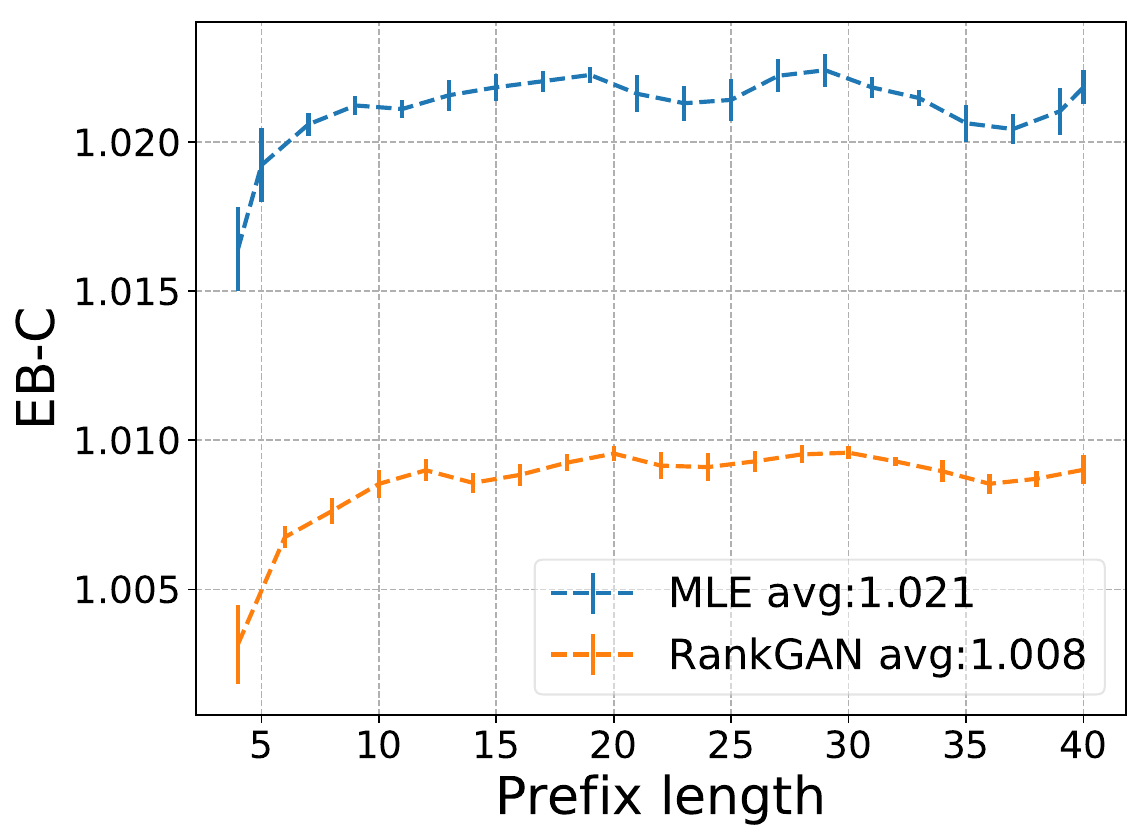}}
  \subfigure[ScratchGAN]{\includegraphics[width=0.49\linewidth]{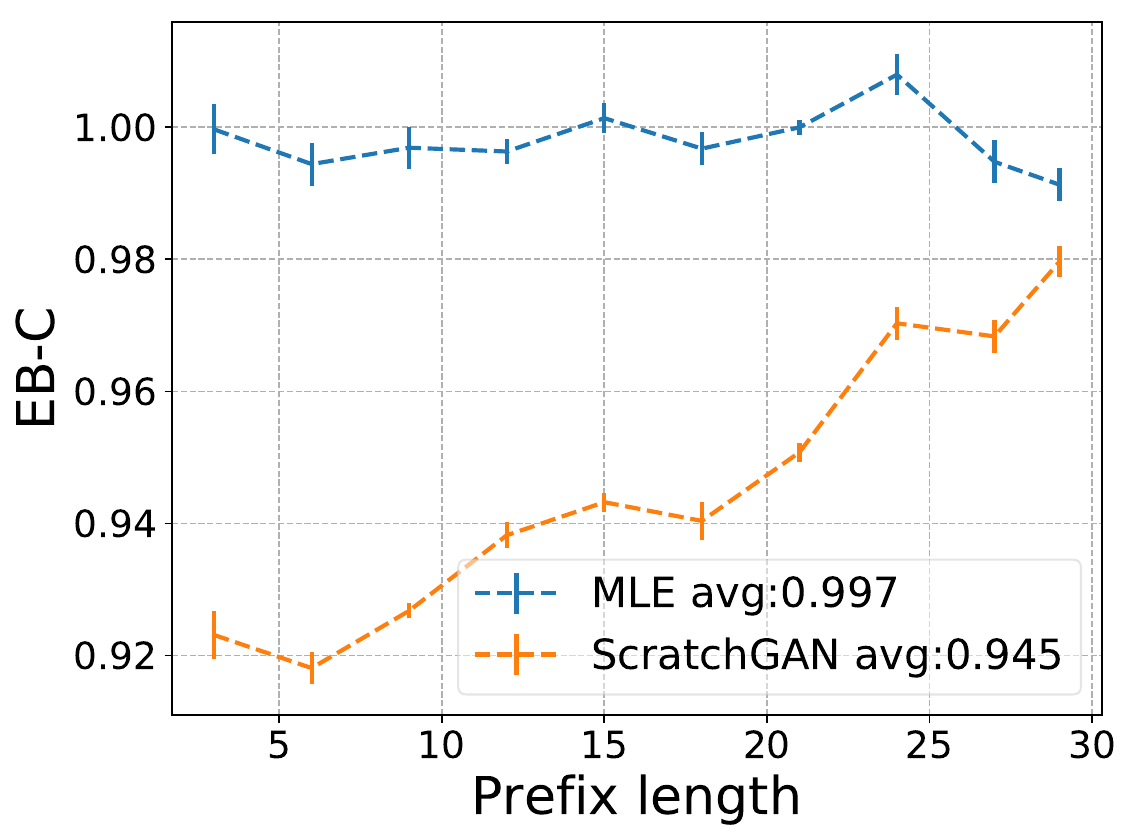}}
  \caption{EB-C measurements (with $d_\text{JS}$) for comparing MLE baseline with text GANs in the synthetic setting. The measurements confirm that the ScratchGAN model behaves better with model prefix than data prefix.}
  \label{fig:ebc_cond_nonMLE_js}
\end{figure}

The results are shown in Figure \ref{fig:ebc_cond_nonMLE_js}.  We find that RankGAN and ScratchGAN give lower EB-C measurements than MLE, which is as expected, as these methods avoid teacher forcing. Most EB-C values in the ScratchGAN case are less than 1, which matches our intuition that GAN models should behave better when fed with model prefixes than data prefixes. On the other hand, EB-C in the RankGAN case is still slightly larger than 1. We believe the reason is that RankGAN still relies on MLE pre-training.

To the best of our knowledge, this is the first direct empirical evidence showing that non-MLE training could indeed avoid the exposure bias problem in that the model behaves better with model prefix than data prefix. It also suggests that EB-C correctly captures how the training-testing discrepancy affects generation. Note that lower EB-C value does not mean the generation performance is better (the authors of ScratchGAN acknowledge their performance is still inferior to the MLE baseline).

\section{Discussions}
\label{app:sec_discuss}

We discuss the critical question ``Is teacher forcing really a \textit{biased} objective?'', from the perspective of objective functions.  Note that the teacher forcing (MLE) objective~(\ref{eq:mleobj}) can be re-written as:
\begin{align}
\small
\begin{split}
    & \argmin_{\theta}\mathop{\mathbb{E}}_{W\sim P_D}-\Sigma_{l=0}^{L-1}\log P_{\theta}(W_{l+1}|W_{1:l})  \\
    = & \argmin_{\theta}\mathop{\mathbb{E}}_{W \sim P_D}-\log P_{\theta}(W) \\
     = & \argmin_{\theta}\mathop{\mathbb{E}}_{W \sim P_D}\log\tfrac{P_D(W)}{ P_{\theta}(W)} \\
    = & \argmin_{\theta} D_\text{KL}(P_D||P_\theta),
\end{split}
\end{align}
where $D_\text{KL}$ denotes the Kullback-Leibler divergence, and $\theta$ denotes the trainable parameters in $P_M$. Therefore, teacher forcing (MLE) training is minimizing the divergence of $P_\theta$, which is exactly the model's sampling distribution, from $P_D$. While it is true that the training is ``exposed'' to data samples as prefixes, we should not simply deduce the objective is ``biased''.

\paragraph{A Concrete Toy Example for EB-C} What kind of model has large EB-C values? Here we discuss a concrete toy LM  which is hand-crafted to have a large EB-C value. However, we will argue that this model is unlikely to be a product of MLE training. 
\begin{example}
\label{eg:cond}
Suppose $L=2$, $V=\{A,B\}$, and the ground-truth data distribution is uniform on $\{AA,AB,BB,BA\}$. $P_M$ is crafted as follows: $P_M(W_1=A)=0.9,P_M(W_2=A|W_1=A)=0.9,P_M(W_2=A|W_1=B)=0.5$. Note that the model behaves worse when $W_1=A$, which is of high probability during sampling.
\end{example}

For Example \ref{eg:cond}, we can easily get that for $d_\text{TV}$, $\text{CGD}(M|D(1))=0.2$ and $\text{CGD}(M|M(1))=0.36$, which gives us $\text{EB-C}(M,1)=1.8$. However, this crafted model is unlikely to be an outcome of MLE training. The fact that $P_M(\cdot\ |\ W_1=B)$ is better modeled suggests that in the training data, there are more sentences beginning with $W_1=B$ than $W_1=A$. So MLE training should assign more probability to $P_M(W_1=B)$ than $P_M(W_1=A)$, not the other way around. From this perspective, the claim of exposure bias seems to be conflicting with the MLE principle.

\section{A Preliminary Study for Machine Translation}
\label{appsec:selfrecover_mt}

\begin{table*}
\vskip 0.15in
\small
\begin{center}
\begin{tabular}{l}
\hline
SOURCE:  sobald der richter mich sah , \\
REF:  and as soon as i walked inside , the judge saw\\ me coming in . \\
DATA3: \underline{and as soon} as the judge saw me . \\
NORMAL: as soon as the judge saw me . \\
UNREL3: \underline{what else is} it that the judge saw me ? \\
RAND3: \underline{still take open} action as the judge saw me . \\
\hline
SOURCE:  ich fuhr also zum gericht . \\
REF:  and i got in my car and i went to this courthouse . \\
DATA3:  \underline{and i got} to the court . \\
NORMAL: so i went to the court . \\
UNREL3:  \underline{the reasons for} me to go to the court . \\
RAND3:  \underline{ge bor last} year , i went to court . \\
\hline
SOURCE:  ich bekam etwas angst vor technologie . \\
REF:  i found myself becoming a little bit of a technophobe . \\
DATA3: \underline{i found myself} a little scared of technology . \\
NORMAL: i got a little scared of technology . \\
UNREL3: \underline{um , my} fear of technology was with me . \\
RAND3: \underline{kids - ds} i got a little scared of technology . \\
\hline
SOURCE: das werde ich ihnen jetzt zeigen \\
REF:  so i 'm going to try and show you what you really \\ get for 10 billion pixels . \\
DATA3: \underline{so i 'm} going to show you this now . \\
NORMAL: this is what i 'm going to show you . \\
UNREL3:  \underline{why did i} show you that now ? \\
RAND3: \underline{told ct happening} to you now . \\
\hline
\end{tabular}
\normalsize
\vspace{0.2cm}
\caption{A standard NMT transformer model fed with different types of length-3 prefix. We did not do any cherry picking. ``DATA'' means the first three output tokens are forced to be from the reference. ``NORMAL'' means no prefix is forced during decoding. ``UNREL'' means the first three tokens are forced to be from another random unrelated sentence (which is wrong but grammatical). ``RAND'' means the first three tokens are completely random words. The given prefixes are underlined. The self-recovery ability is also observed in this setting.}
\label{tab:nmt_eb}
\end{center}
\vskip -0.1in
\end{table*}

In this section, we conduct a preliminary prefix-switching experiment for a standard neural machine translation (NMT) setting. 

We follow the example code from Fairseq\footnote{\url{https://github.com/pytorch/fairseq/tree/master/examples/translation}}, to train a 6-layer encoder-decoder transformer model with a hidden dimension of 512 and an inner dimension of 1024 on the IWSLT14 German-to-English dataset.\footnote{\url{http://workshop2014.iwslt.org/}} It has around 160k sentences / 3.7m words for training, and 6.7k sentences / 150k words for validation or testing (in English). For decoding we use beam-search with beam 20.

We feed the trained model with different types of prefixes during decoding which represents different levels of training-generation discrepancy. The samples are shown in Table \ref{tab:nmt_eb}. Note that the source input is kept intact. 

Note that we should not directly compare the generation with data or model prefix with the corresponding reference, because in a constrained task such as MT, it would be too much cheating to give partial data reference to the model. Still, we observe that data prefixes do not greatly improve the generation. But, it could also be due to the short generation length, and errors from exposure bias did not build up yet.

More interestingly, in the extreme case of unrelated or random prefix, the model still generates fairly good partial translation. This suggests the self-recovery ability does not only exist for LMs trained for open-ended generation. Finally, we emphasize that the existence of self-recovery does not rule out the possibility that exposure bias could still be serious for machine translation. A comprehensive and principled study is needed (preferably with datasets of longer sequences) and we leave that as future work.


\end{document}